\documentclass[11pt]{article}

\usepackage[truedimen,margin=25truemm]{geometry}

\usepackage{graphicx} 
\usepackage{epsfig} 
\usepackage{subfigure} 

\usepackage{amsmath}
\usepackage{amsthm}
\usepackage{amssymb}

\usepackage{algorithm}
\usepackage{algorithmic}

\newtheorem{definition}{Definition}

\usepackage{color}

\usepackage{ulem}

\usepackage{booktabs}

\usepackage{hyperref}

\usepackage{graphicx} 
\usepackage{epsfig}
\usepackage{subfigure}

\usepackage{amssymb}
\usepackage{amsmath}
\usepackage{amsthm}
\usepackage{mathtools}
\usepackage[raggedright]{titlesec}

\graphicspath{{./figure/}}

\def\vec#1{\mbox{\boldmath $#1$}}

\title{Graph embedding using multi-layer adjacent point merging model}

\author{Jianming Huang \thanks{Graduate School of Fundamental Science and Engineering, WASEDA University, 3-4-1 Okubo, Shinjuku-ku, Tokyo 169-8555, Japan (e-mail: koukenmei@toki.waseda.jp) } \and Hiroyuki Kasai \thanks{Department of Computer Science and Communication Engineering, WASEDA University, 3-4-1 Okubo, Shinjuku-ku, Tokyo 169-8555, Japan (e-mail: hiroyuki.kasai@waseda.jp)}}

\begin{document}

\maketitle

\begin{abstract}
For graph classification tasks, many traditional kernel methods focus on measuring the similarity between graphs. These methods have achieved great success on resolving graph isomorphism problems. However, in some classification problems, the graph class depends on not only the topological similarity of the whole graph, but also constituent subgraph patterns. To this end, we propose a novel graph embedding method using a multi-layer adjacent point merging model. This embedding method allows us to extract different subgraph patterns from train-data. Then we present a flexible loss function for feature selection which enhances the robustness of our method for different classification problems. Finally, numerical evaluations demonstrate that our proposed method outperforms many state-of-the-art methods.\\
\end{abstract}

\begin{center}
{\small
Published in\\
IEEE International Conference on Acoustics, Speech and Signal Processing (ICASSP) 2021 \cite{huang_icassp_2021}.
}
\end{center}

\section{Introduction}
Graph-structured data have been used widely in various fields, such as chemoinformatics, bioinformatics, social networks, and computer vision \cite{vishwanathan2010graph_s,kriege2020survey,hashimoto_ICASSSP_2020,horie_EUSIPCO_2020}. In graph-structured data classification tasks, many efforts have been done to define a measure between graphs, some of which are well-known as {\it graph kernels}. The graph kernels have been used widely for several decades, and are still developing rapidly in recent years. The graph kernels are kernel functions that compute similarity between two graphs, and have shown effective performances in graph classification tasks using machine learning algorithms. However, these graph kernel are not specifically focusing on the classification problem itself, but are focusing on resolving the graph and subgraph {\it isomorphism problem}, which aims to evaluate how two graphs are similar. In some graph classification problems, the graph class may not only depend on the topological similarity of the whole graph, but also subgraph patterns which is equally important. Furthermore, the key feature of subgraphs that have big impacts on classification performances are different in each individual classification problem. Nevertheless, many of these existing methods seem to neglect these points. To this end, we propose a multi-layer adjacent point pattern embedding, which can extract and select effective subgraph patterns from graphs automatically for different classification problem.  Our contributions can be summarized as described below:
\begin{itemize}
	\item We present a {\it multi-layer adjacent point merging} model, which can extract multi-granularity representations of subgraphs, i.e., simple-to-complex subgraphs.
	\item We propose a flexible loss function for feature selection, which makes our proposed method robust to different key features required in each classification problem.
\end{itemize}

Hereinafter, we represent scalars as lower-case letters $(a, b, \ldots)$, and vectors as bold typeface lower-case letters $(\vec{a}, \vec{b}, \ldots)$.
We write $\mathbb{R}^{{n}}$ to denote $n$-dimensional vector. 
A graph is a pair $G=(\mathcal{V}, \mathcal{E})$ consisting of a set of $n$ vertices (or nodes) $\mathcal{V}=\{v_1, v_2, \ldots, v_n\}$ and a set of $m$ edges $\mathcal{E} \subseteq \mathcal{V} \times \mathcal{V}$. $G$ is an undirected graph if a graph $G$ includes only edges with no direction. 
The numbers of vertices and edges are, respectively, $|\mathcal{V}|$ and $|\mathcal{E}|$. If two vertices, say $v_i, v_j \in \mathcal{V}$, are connected by an edge $e$, then this edge is denoted as $e_{ij}$. These two vertices are said to be adjacent or neighbors. We consider only undirected graphs with no self-loop.

\section{Related Work}
The graph classification has developed for several decades. To compute similarity between graphs in various data mining tasks, random walk kernel \cite{gartner2003graph} has been developed 
for graph classification. This method is based on the counting of matching random walks in two graphs with a label or not. However, random walk kernel faces a difficulty by which the computational cost is $O(n^6)$ for comparing a pair of graphs in graph product space, which is a non-negligible cost, especially for large-scale graphs. To resolve this difficulty, 
subsequent work on Weisfeiler--Lehman graph kernel \cite{Shervashidze_JMLR_2011_s} has brought great success. They improved the original Weisfeiler--Lehman test using a form of multiple iteration, where neighbor patterns are aggregated. 

In recent years, as effective performances of optimal transport theory \cite{Villani_2008_OTBook_s, Peyre_2019_OTBook_s} in a machine learning domain \cite{Kasai_ICASSP_2020}, graph kernel methods are also improved greatly when combined with optimal transport theory \cite{Huang_arXiv_LCS_2020}. Recent research by \cite{togninalli2019wasserstein_s}, presents a Wasserstein-based Weisfeiler--Lehman graph kernel (WWL), which maps node embedding of a Weisfeiler--Lehman pattern to a feature space, and which computes kernel values using the Wasserstein distance of two point clouds in the feature space. They received better results than those yielded by the original Weisfeiler--Lehman kernel. GOT \cite{maretic2019got_s} uses optimal transport differently to compute the Wasserstein distance between two normal distributions derived by graph Laplacian matrices, instead of generating walks or comparing vertex neighbors in graphs. 
Another attractive work by \cite{titouan2019optimal_s} raises difficulties that both the Wasserstein distance and the Gromov--Wasserstein distance are unable to accommodate the graph structure and feature information. To resolve this difficulty, they propose a notion of Fused Gromov--Wasserstein (FGW) distance, which considers both structure characteristics and feature information of two graphs.

\section{Multi-layer Adjacent Point Merging}
\subsection{Adjacent Point Merging}
\subsubsection{Adjacent Point Pattern (APP)}
\begin{figure}[t]
\centering
\includegraphics[width=0.5\textwidth]{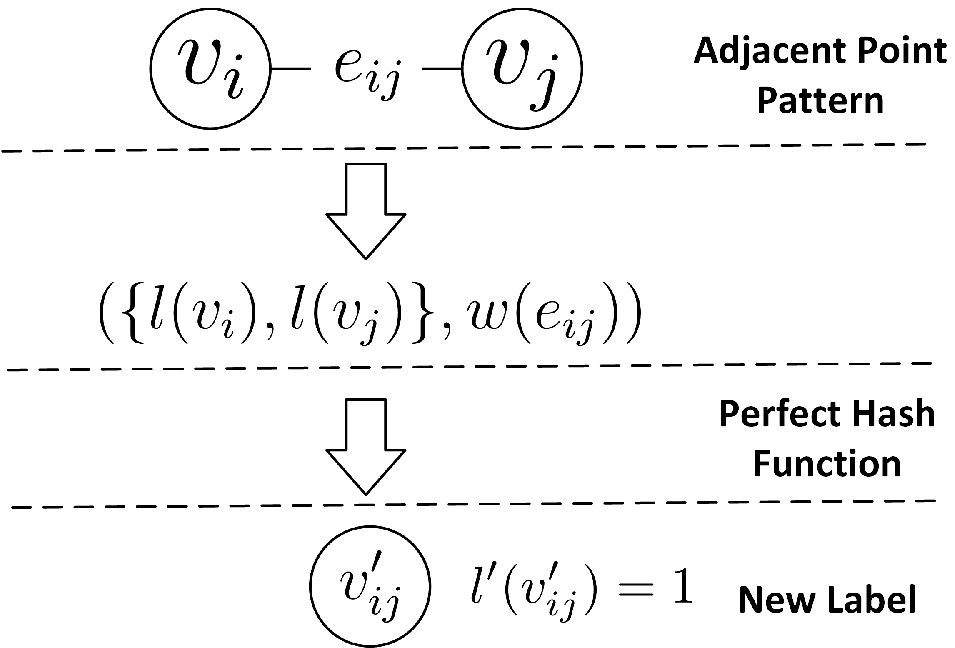} 
\caption{The structure of an APP, which consists of two directly connected vertices $v_i, v_j$ and their connecting edge $e_{ij}$. The APP do not care about the order of $v_i, v_j$, so $(v_i, e_{ij}, v_j)$ and $(v_j, e_{ij}, v_i)$ are equivalent. An APP is finally merged as a vertex, and relabeled in the APM.}
\label{figure:APP} 
\vspace*{0.5cm}

\centering
\includegraphics[width=0.4\textwidth]{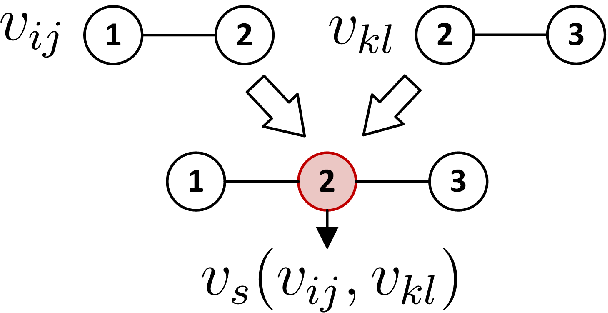} 
\caption{The sharing point of two APPs $v_s(v_{ij}, v_{kl})$. }
\label{figure:sp} 
\end{figure}

To extract subgraph features from a graph, we start from a simple pattern of subgraph which is called {\it adjacent point pattern} (APP). The APP is composed of two part: (1) two vertices which are directly connected with each other; (2) the connecting edge between these vertices. As shown in Figure \ref{figure:APP}, $v_i$ and $v_j$ denote  two adjacent points, and $e_{ij}$ denotes their connecting edge. When both vertices and edge are assigned with discrete labels, there could be different APPs with different permutations of labels of $v_i, v_j$ and $e_{ij}$. Therefore, we exploit a {\it perfect hash method} \cite{cichelli1980minimal} to assign each APP with a distinguishable label. We propose the definition of APP as:

\begin{definition}[Adjacent Point Pattern]
Given two vertices $v_i, v_j$ and the connecting edge $e_{ij}$ between $v_i$ and $v_j$. Let $l : \mathcal{V} \to \Sigma$ denote a function that maps a vertex object v to its categorical node label assigned from a finite label alphabet $\Sigma$, where $\mathcal{V}$ denotes a certain vertex set which $v$ belongs to. Furthermore, let $w : \mathcal{E} \to \Sigma$ be the edge label mapping function, where $\mathcal{E}$ denotes a certain edge set. The adjacent point pattern is defined as 
$F_{\rm APP}(v_i, e_{ij}, v_j) = \left(\{l(v_i), l(v_j)\},w(e_{ij})\right),$
where $F_{\rm APP} : \mathcal{V}\times\mathcal{E}\times\mathcal{V} \to \mathbb{H}$ is the function which maps the adjacent vertices and connecting edge to a Hilbert space $\mathbb{H}$ of the inner production of vertex label set and edge label. In the special condition in which edge $e_{ij}$ has no label, the vertex-only adjacent point pattern is defined as 
$F_{\rm APP}(v_i, v_j) = \left(\{l(v_i), l(v_j)\}\right).$
\end{definition}

For the label assigning, we use a perfect hash function $F_{\rm hash}: \mathbb{H} \to \Sigma$, such that, for two APPs, $\vec{x}_1$ and $\vec{x}_2$, $F_{\rm hash}(\vec{x}_1) = F_{\rm hash}(\vec{x}_2)$ if and only if $\vec{x}_1 = \vec{x}_2$. We set $F_{hash}(\vec{x}_i)$ as the {\it Adjacent Point Pattern Label} of $\vec{x}_i$.

\subsubsection{Adjacent Point Merging (APM)}
To generate APPs in a labeled graph, we introduce an operation of {\it adjacent point merging} (APM). This operation performs merging all pairwises of vertices in a graph, which are directly connected through an edge. 

\begin{definition}[Adjacent Point Merging]
Given an undirected and connected graph $G = (\mathcal{V}, \mathcal{E})$ with a set of vertices $\mathcal{V} = \left\{v_i\right\}^N_{i=1}$
and a set of edges $\mathcal{E} = \left\{e_{ij}\right\}$ such that $|\mathcal{V}| = N$ and $|\mathcal{E}| = M$. Both vertices and edges in $G$ are assigned a categorical label, and $l : \mathcal{V} \to \Sigma$ and $w : \mathcal{E} \to \Sigma$ are the label mapping function of vertex and edge, respectively. 
For each $v_j \in \mathcal{V}$, it has $\mathcal{A}_j = \left\{(v_i, e_{ij}, v_j):v_i, v_j \in \mathcal{V}, v_i \in \mathcal{N}(v_j), e_{ij} \in \mathcal{E}\right\}$, where $\mathcal{N}(v_j)$ denotes the neighborhood of $v_j$ and $e_{ij}$ is the connecting edge between $v_i$ and $v_j$. 
Then, the operation of the adjacent point merging follows steps as:
(1) For each $\mathcal{A}_j$, we create a new vertex set $\mathcal{V}^\prime_j = \left\{v_{ij}^\prime\right\}$, in which a single vertex $v_{ij}^\prime$ represent a pairwise of adjacent vertices $(v_i, e_{ij}, v_j)$ in $\mathcal{A}_j$;
(2) Generating $\mathcal{V}^\prime_j$ for all $j\in [N]$, we obtain $\mathcal{V}^\prime$ by removing identical vertices as $\mathcal{V}^\prime = \bigcap_{j=1}^N \mathcal{V}_j^\prime$.
(3) The new adjacent relationship between new vertices $v_{ij}^\prime \in \mathcal{V}^\prime$ is defined as: $v_{ij}^\prime \in \mathcal{N}(v_{kl}^\prime)$ holds if $\exists v \in \left\{v_i, v_j\right\} \cap \left\{v_k, v_l\right\}$, we call this a sharing point of $v_{ij}$ and $v_{kl}$ and write it $v_s(v_{ij}, v_{kl})$, which is shown in Figure \ref{figure:sp}. Then we have a new edge set $\mathcal{E}^\prime = \left\{e_{ij,kl}^\prime\right\}$ in which edge $e_{ij,kl}^\prime$ connects vertices $v_{ij}^\prime$ and $v_{kl}^\prime$ directly;
(4) Then we define new label mapping functions of vertex and edge as $l^\prime:\mathcal{V}^\prime \to \Sigma$ and $w^\prime : \mathcal{E}^\prime \to \Sigma$, respectively, given by
\begin{eqnarray}
\label{Eq:nlabel}
	l^\prime(v_{ij}^\prime) &=& F_{\rm hash}(F_{\rm APP}(v_i, e_{ij}, v_j)),\cr
	w^\prime(e_{ij,kl}^\prime) &=& l(v_s(v_{ij}, v_{kl})).
\end{eqnarray}
(5) Using these new components, we create a new graph $\!G_A(G)\! =\! (\mathcal{V}^\prime, \mathcal{E}^\prime)\!$ with new label mapping function $l^\prime$ and $w^\prime$. 
\end{definition}

\subsection{Multi-layer Adjacent Point Merging}
\begin{figure}[t]
\centering
\includegraphics[width=0.7\textwidth]{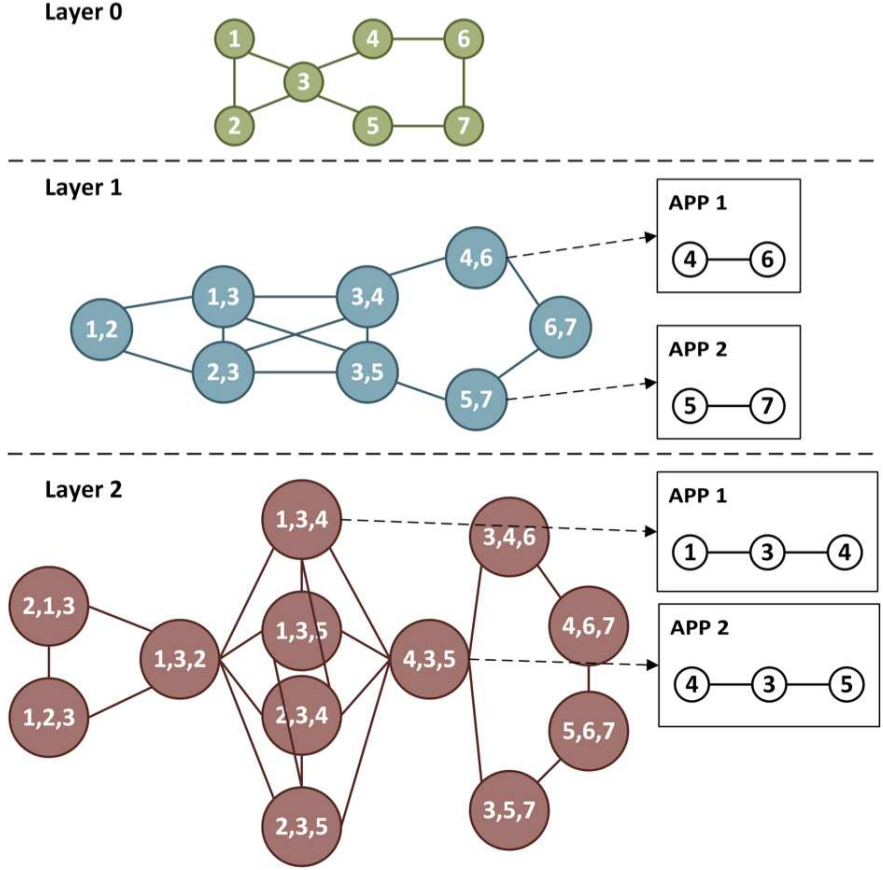} 
\caption{The structure of a 2-layer Adjacent Point Merging. The numbers in the nodes denotes the vertex index of input graph (layer 0). The graphs in each layer is the result of performing APM on the previous layer's graph. Any vertex in $i$-th layer ($i \geq 1$) represent a class of APP in input graph.}
\label{figure:MLAPM} 
\end{figure}

Through the operation of APM elaborated in previous section, we are able to extract and relabel a very simple subgraph pattern. However, we find that this operation is iteratively workable. Therefore, we introduce a multi-layer structure of the {\it adjacent point merging}. The benefit of this multi-layer structure is to provide multi-granularity representations of subgraphs, i.e., simple-to-complex subgraphs, of an entire graph $G$.
More specifically, given an undirected and connected graph $G = (\mathcal{V}, \mathcal{E})$ with categorical label assigned to vertex and edge, one iteration of the {\it multi-layer adjacent point merging} is simply written as $G_{i+1} = G_A(G_i)$,
where $G_i$ is the transformed graph after $i$ times of APM. Specially, $G_0$ is equal to the original graph $G$. As shown in Figure \ref{figure:MLAPM}, there is an example of {\it 2-layer} adjacent point merging, which is able to extract a 3-vertices subgraph pattern. 

\section{Adjacent Point Pattern Embedding}
\subsection{Embedding using All Adjacent Point Pattern Label}
Based on the multi-layer APM structure, we propose the {\it adjacent point pattern embedding} (APPE) method to generate graph embedding. In a $K$-layer APM, we can simply use the number of vertices which have the same label in the top-layer transformed graph $G_K$ as element of our embedding. In this case, given a graph $G=\{\mathcal{V}, \mathcal{E}\}$ with $l(\cdot)$ and $w(\cdot)$ as the label mapping function of vertex and edge, respectively, we compute graph embedding through the following steps: (1) We use the $K$-layer APM to transform $G$. In the $i$-th layer, we obtain a transformed $G_i = (\mathcal{V}_i,\mathcal{E}_i)$ and new label sets $\Sigma_\mathcal{V}^i, \Sigma_\mathcal{E}^i$ of vertex and edge (Equation \ref{Eq:nlabel}); (2) We use the output of the $K$-th layer to compute the final embedding. For each discrete label $L_j \in \Sigma_\mathcal{V}^K$, which represent a certain subgraph pattern in the original graph $G$, we count all the vertices with the same label as $L_j$ in graph $G_K$. Then we obtain a vertex set $\mathcal{L}_j = \left\{v:l^\prime(v) = L_j, v \in \mathcal{V}_K \right\}$, where $l^\prime:\mathcal{V}_K \to \Sigma_\mathcal{V}^K$ denotes the vertex label mapping function of $G_K$. Using all the $\mathcal{L}_j$, we compute our APPE of graph $G$ as:
\begin{equation}
	\label{Eq:embed1}
	\vec{e}(G) = [ |\mathcal{L}_1|, |\mathcal{L}_2|, ..., |\mathcal{L}_{|\Sigma_\mathcal{V}^K|}|] \in \mathbb{R}^{|\Sigma_\mathcal{V}^K|}.
\end{equation}

\subsection{Feature Selection}

As the number of layer increases, the size of top-layer vertex label set $\Sigma_\mathcal{V}^K$ may become too large, which makes the size of embedding become too large. To prevent this, we perform a feature selection strategy, and only use a part of dimensions as our embedding. In order to enhance the robustness of our method, we propose a new loss function to evaluate how effective each feature is for the classification task.

For $C$-classification problem with $K$-layer APM, we maintain $C$ weight vectors as $\{\vec{w}_i\in \mathbb{R}^{|\Sigma_\mathcal{V}^K|}\}_{i=1}^{C}$, where $|\Sigma_\mathcal{V}^K|$ is the size of $\vec{e}(G)$ in Equation (\ref{Eq:embed1}). Each weight vector $\vec{w}_i$ corresponds to the $i$-th graph class, and each dimension in $\vec{w}_i$ corresponds to a dimension of $\vec{e}(G)$. Given the train data graph set $\mathcal{G}$ and the class label mapping function $y: \mathcal{G} \to \Sigma_{\mathcal{G}}$ which maps a graph in $\mathcal{G}$ to an integer class label $\Sigma_{\mathcal{G}}$, we perform the feature selection following the steps: 

\noindent 
(Step.1) We first update the weight vectors $\vec{w}_i$ by the formula: $\vec{w}_i = \sum_{G \in \mathcal{G}, y(G)=i} \vec{e}(G)$.

\noindent 
(Step.2) After updating all of the $\vec{w}_i$, we compute the final loss of each dimension $\vec{w}_{\rm loss}$ as:
\begin{eqnarray}
	\label{Eq:loss}
	\vec{w}_{\rm loss} &=& \max \left( F_{\rm loss}(1), F_{\rm loss}(2), ..., F_{\rm loss}(C) \right),
\end{eqnarray}
where the $j$-th dimension of $\vec{w}_{\rm loss}$ represents the loss of the $j$-th dimension for the the $i$-th graph class, and $F_{\rm loss}(i)$ is:
$F_{\rm loss}(i) = -\frac{(\vec{w}_i - \sum_{j =1, j\neq i}^C\vec{w}_j)^2}{\sum_{j = 1}^C \vec{w}_j}.$
It should be noted that all the calculation in the formula is element-wise calculation. In $F_{\rm loss}(i)$, the first term of the numerator denotes the weight of each dimension for the $i$-th graph class. And the second term of the numerator represents the weight of each dimension for graph class except $i$. Therefore, the value of numerator exactly evaluates how effective each dimension is for classifying the $i$-th graph class. The denominator in the formula is used for normalization. We also compute the square of the numerator in order to prevent the situation where some dimensions which has too small weight may obtain a small loss value. For example, assume that $\vec{e}(G)$ only has one dimension, comparing the case with $\vec{w}_i = [1]$ and $\sum_{j =1, j\neq i}^C\vec{w}_j = [0]$, to the case with $\vec{w}_i = [100]$ and $\sum_{j =1, j\neq i}^C\vec{w}_j = [1]$, the former gets a smaller $F_{\rm loss}(i)$ than the latter does without squaring the numerator. This is unreasonable because in the second case the dimension weights more than the first in total of $\sum_{i=1}^C\vec{w}_i$. 

\noindent 
(Step.3) We finally select the top-$D$ dimensions of $\vec{e}(G)$ which have the least loss in $\vec{w}_{\rm loss}$ as a final embedding:
$\vec{e}_{\rm fin}(G) = \left[ \vec{e}(G)(j): \vec{w}_{loss}(j) \leq {\rm sort}(\vec{w}_{loss})(D) \right] \in \mathbb{R}^D$,
where ${\rm sort}(\cdot)$ denotes sorting a vector in ascending order.

\subsection{Computational Complexity}
We present the analysis of computational complexity in this section. For 1-layer APM, it is obvious that the number of vertices in the first-layer transformed graph $G_1$ is equal to the number of edges $|\mathcal{E}|$ in $G$. Therefore, the computational complexity for a single graph $G(\mathcal{V},\mathcal{E})$ is $O(|\mathcal{E}|)$. For 2-layer APM, the computational complexity is $O(\sum{D^2}-|\mathcal{E}|)$, where $\sum{D^2}$ denotes the sum of the square of the degree of each vertex, which is based on the formulation of computing number of edges of the line graph \cite{younger1972graph}.

\begin{table*}
\centering
\caption{Average classification accuracy on graph datasets ($D=100$)}
\label{Tab:acc1}
\begin{tabular}{lccccc}
\toprule
M{\scriptsize ETHOD}&MUTAG&BZR&COX2&PROTEINS&Mutagenicity\cr
\midrule
GK \cite{shervashidze2009efficient_s}&86.19$\pm$6.68&79.77$\pm$2.48&78.16$\pm$1.17&72.05$\pm$4.09&59.92$\pm$1.77\cr
RWK \cite{gartner2003graph} &84.53$\pm$7.79&78.53$\pm$1.38&80.29$\pm$2.16&-&69.03$\pm$1.93\cr
WWL \cite{togninalli2019wasserstein_s} &79.70$\pm$6.91&\textbf{88.39$\pm$3.88}&79.65$\pm$4.50&74.03$\pm$5.01&80.86$\pm$2.25\cr
FGW \cite{titouan2019optimal_s} &80.26$\pm$9.47&83.95$\pm$2.45&78.15$\pm$2.12&71.15$\pm$3.63&67.87$\pm$1.70\cr
AWE \cite{ivanov2018anonymous} &83.56$\pm$3.50&82.45$\pm$4.66&79.00$\pm$3.88&67.38$\pm$2.75&73.76$\pm$1.96\cr
\midrule
1-layer APM (proposed)&87.74$\pm$4.79&86.64$\pm$3.06&80.09$\pm$2.80&\textbf{75.38$\pm$4.61}&77.91$\pm$1.49\cr
2-layer APM (proposed)&\textbf{88.80$\pm$5.57}&85.16$\pm$3.21&\textbf{82.01$\pm$2.16}&74.12$\pm$4.84&\textbf{81.50$\pm$1.66}\cr
\bottomrule
\end{tabular}
\end{table*}

\begin{table}
\centering
\caption{Accuracy under different loss computings ($D=10$)}
\label{Tab:acc2}
\begin{tabular}{lccc}
\toprule
M{\scriptsize ETHOD}&BZR&Mutagenicity\cr
\midrule
2-layer APM&\textbf{86.65$\pm$4.34}&\textbf{75.92$\pm$1.94}\cr
2-layer APM (mean)&85.90$\pm$3.00&74.59$\pm$1.13\cr
\bottomrule
\end{tabular}
\end{table}

\section{Numerical Evaluation}
For numerical experiments, we evaluate the 1-layer and 2-layer APM. For feature selection, we set up a maximal dimension $D$ as $100$.
We compare our proposed methods to five state-of-the-art methods as follows: (1) Traditional methods: the Graphlet Kernel (GK) \cite{shervashidze2009efficient_s} and Random Walk Kernel (RWK) \cite{gartner2003graph}. For implementation of all the $\mathcal{R}$-convolution kernels, we use the Grakel python library \cite{Siglidis_JMLR_2020_s}. (2) Recent year's methods: Wasserstein Weisfeiler--Lehman Kernel (WWL) \cite{togninalli2019wasserstein_s}, Fused Gromov--Wasserstein Kernel (FGW) \cite{titouan2019optimal_s} and
 Anonymous Walk Embeddings (AWE) \cite{ivanov2018anonymous}. The parameter of $H$ in WWL is set as $4$. The shortest path matrix is used in FGW. The stepsize of AWE is set to $3$.
For classification, we train a multi-class SVM classifier using one-vs.-one approach, and apply $10$ times of nested cross-validation with a 10-fold inner cross-validation. For parameter adjustment of SVM, we apply a grid search with SVM parameter $C$ within $\{ 0.001, 0.01, 0.1, 1, 10, 100, 1000\}$. Then we calculate an average accuracy and a standard deviation after classification. 

We use several widely used benchmark real-world datasets. Among them, MUTAG \cite{debnath1991structure_s}, Mutagenicity \cite{riesen2008iam_s} include graphs with discrete labels of both vertices and edges. BZR, COX2 \cite{sutherland2003spline_s}, and PROTEINS \cite{borgwardt2005protein} include graphs with discrete and continuous vertex attributes. All datasets above are available in TUD dataset \cite{Morris+2020}.
Because our proposed method only supports graphs with discrete labels of vertices and/or edges, we remove continuous attributes from original graphs, making sure that only discrete labels remain.
Table \ref{Tab:acc1} shows the average classification accuracies on different graph datasets, where the top accuracy in each dataset is in bold. The results marked with ``-'' indicate that dataset is not applicable for objective methods.
Overall, our proposed methods are shown to outperform many state-of-the-art methods in all the datasets except BZR. However, our methods still give the second-best results next to WWL. Table \ref{Tab:acc2} is a condition-controlled experiment where we use a mean weight $\vec{w}_{\rm loss} = \frac{1}{C}\sum_{i=1}^C \vec{w}_i$ to replace Equation (\ref{Eq:loss}), which shows the effectiveness of our loss computing. It should be noted that if $D$ is too large, the final embedding will include excessive dimensions so that it is hard to tell the difference between loss computings. Therefore, $D$ is set to $10$ in this experiment.

\section{Conclusion}
We have proposed a new method of computing graph embedding using a {\it multi-layer adjacent point merging} model, which extracts subgraph patterns and utilizes a flexible loss function to select effective ones of them. The numerical experiments revealed that our proposed methods have better performances than those of others. As future work, we specifically expect a more light-weight way to extracts subgraphs in high-layer models because the complexity of the current model increases rapidly as the number of layers increases.

\clearpage	
\bibliographystyle{unsrt}
\bibliography{strings,refs,graph_datasets,graph,optimal_transport}

\end{document}